\documentclass[letterpaper]{article} 
\usepackage{aaai2027}  
\usepackage[hyphens]{url}  
\usepackage{graphicx} 
\usepackage{natbib}  
\usepackage{caption} 
\usepackage{algorithm}
\usepackage{algorithmic}

\usepackage{newfloat}
\usepackage{listings}
\DeclareCaptionStyle{ruled}{labelfont=normalfont,labelsep=colon,strut=off} 
\floatstyle{ruled}
\newfloat{listing}{tb}{lst}{}
\floatname{listing}{Listing}

\usepackage{booktabs}

\usepackage{amsmath}
\usepackage{amssymb}
\usepackage{pifont} 
\usepackage{xcolor} 
\usepackage{colortbl} 

\title{CodecArena: Codec Quality Assessment via Visual Reinforcement Learning}
\author{
    Jiaye Fu\textsuperscript{\rm 1,\rm 2}\setcounter{footnote}{1}\thanks{These authors contributed equally.},
    Weiqi Li\textsuperscript{\rm 1,\rm 3}\footnotemark[2],
    Qiankun Gao\textsuperscript{\rm 1},
    Yanchen Zhao\textsuperscript{\rm 2},\\
    Xiandong Meng\textsuperscript{\rm 3},
    Jian Zhang\textsuperscript{\rm 1},
    Siwei Ma\textsuperscript{\rm 2},
    Jiaqi Zhang\textsuperscript{\rm 2}\setcounter{footnote}{0}\thanks{Corresponding author.}
}
\affiliations{
    \textsuperscript{\rm 1}School of Electronic and Computer Engineering, Peking University\\
    \textsuperscript{\rm 2}School of Computer Science, Peking University\\
    \textsuperscript{\rm 3}Pengcheng Laboratory
}

\begin{document}

\maketitle


\begin{abstract}
Video coding is advancing into the low and ultra-low bitrate regime, driven by end-to-end codecs that replace the hand-crafted pipeline with jointly optimized neural networks and generative codecs that exploit the priors of video generation models. Yet the dominant metrics, LPIPS and DISTS, measure feature and texture similarity rather than content fidelity: a reconstruction that hallucinates a wrong face or blurs text into convincing strokes can still score well, even when a human rejects it instantly. To address this, we propose \textbf{CodecArena}, the first vision-language framework for video coding quality assessment, casting codec evaluation as source-conditioned comparative reasoning between a reference and its reconstructions. We optimize CodecArena with \textbf{Facet-GRPO}, a visual reinforcement learning scheme that aligns pairwise codec preferences while grounding the verdict in five fidelity facets: identity, objects, text, texture, and temporal consistency. Its facet-anchored reward uses automatically derived facet directions as weak anchors, rather than human per-facet labels, to prevent any single sub-score from dominating the holistic preference and to yield interpretable fine-grained quality judgments. To support training and evaluation in this underexplored regime, we construct two complementary resources: CodecArena-1K, a fully automatic preference dataset of 1,500 comparison groups built from traditional, neural, and generative codec reconstructions with fused vision-language and objective supervision; and CodecArena-Bench, a human-ranked benchmark with source-disjoint videos for fair out-of-domain evaluation. Extensive experiments demonstrate that CodecArena achieves state-of-the-art agreement with human judgments on source-disjoint content across diverse codecs and bitrates, surpassing perceptual metrics and prior vision-language evaluators. The project page is: \url{https://jyfu-vcl.github.io/codecarena}.
\end{abstract}

\section{Introduction}

Video coding is central to streaming, communication, and storage, and its rate-distortion efficiency has been refined for decades. End-to-end neural codecs \citep{li2023neural,jia2025towards} and, more recently, generative codecs \citep{chen2025generative,qi2025generative,mao2025generative} now reach into the low and ultra-low bitrate regime, where fine content is no longer transmitted but synthesized by the decoder. These reconstructions stay sharp where conventional codecs break down, but the distortion changes in kind: blocking and blur give way to hallucinated detail that looks plausible yet departs from the source.

This regime undermines the metrics on which codec research relies. LPIPS \citep{zhang2018unreasonable} and DISTS \citep{ding2022image}, the perceptual metrics most widely adopted to report progress in recent generative coding \citep{chen2025generative,qi2025generative,xue2025s2vc,zheng2026gvc1d}, average feature distances over the whole frame; a globally realistic reconstruction therefore scores well even when a face takes on a different identity, text becomes illegible, or texture is replaced by convincing but incorrect detail, the failures a viewer notices first. This bias is self-reinforcing: many generative codecs are themselves trained against an LPIPS-style perceptual loss \citep{mentzer2021towards,qi2025generative,mao2025generative}, so evaluating with the same metric rewards reconstructions that exploit it rather than preserve content. Because these metrics operate frame by frame, they are also blind to temporal artifacts such as flicker and identity drift.

\begin{figure}[!t]
    \centering
    \includegraphics[width=\columnwidth]{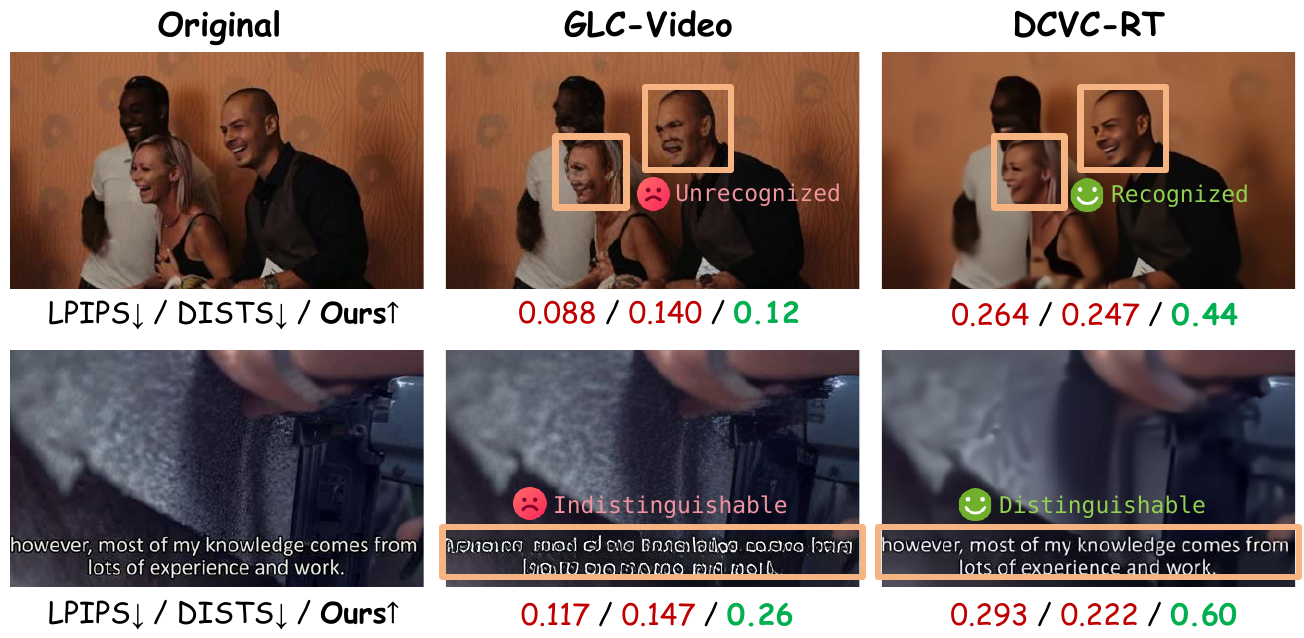}
    \caption{At a matched bitrate of around $0.15$ bpp, LPIPS and DISTS rank GLC-Video above DCVC-RT, yet GLC-Video corrupts the face identity (top) and on-screen text (bottom) that DCVC-RT preserves; CodecArena instead prefers DCVC-RT, matching human judgment. ($\downarrow$/$\uparrow$: lower/higher is better.)}
    \label{fig:badcase}
\end{figure}

Judging whether a reconstruction is faithful is therefore a question of content rather than signal distance. It requires recognizing that a face still belongs to the same person, reading whether on-screen text is unchanged, and checking that objects, textures, and motion remain consistent with the reference. These are the perceptual, world-knowledge, and reasoning abilities of vision-language models, and they mirror how a person rates a codec: by inspecting the source beside its reconstructions and weighing what is preserved. Casting codec quality assessment as vision-language reasoning is thus the natural formulation of the task rather than a borrowed technique. Reinforcement-trained vision-language evaluators already reason reliably about natural and generated content \citep{li2025qinsight,zhang2025vqinsight}, yet none targets the comparative, source-conditioned coding setting, in which reconstructions of one source from heterogeneous codecs must be ranked by faithfulness. Quality assessment for video coding therefore remains open.

We address this with \textbf{CodecArena}, a vision-language judge that, conditioned on the source video, rates how faithfully each reconstruction preserves identity, objects, text, texture, and temporal consistency, and ranks competing reconstructions against one another. The central difficulty is that human preference supervision is naturally holistic: annotators can reliably rank which reconstruction is more faithful, but obtaining calibrated human labels for every quality facet is costly and ambiguous. Faithful codec evaluation, however, must still reason about each facet separately. To bridge this gap, we propose \textbf{Facet-GRPO}, a group-relative policy optimization scheme that learns from holistic preferences while using automatically derived facet directions as weak anchors. By tying the verdict to multiple fidelity facets, Facet-GRPO prevents any single sub-score from dominating the holistic judgment and enables calibrated, fine-grained scoring without manual per-facet annotation.

Supervision of this form does not yet exist, and we therefore construct it. Our pipeline re-encodes natural source videos with traditional, neural, and generative codecs across a range of bitrates, gathers the reconstructions of each source into comparison groups, and labels them automatically by fusing vision-language pairwise judgments with objective perceptual signals under a within-codec rate-monotonicity constraint, producing \textbf{CodecArena-1K}, a corpus of 1{,}500 preference groups built without manual annotation. To measure progress where conventional metrics fail, we further release \textbf{CodecArena-Bench}, the first quality-assessment benchmark for low and ultra-low bitrate coding degradations, with 80 source videos spanning diverse scenes, resolutions, and codecs, labeled by weighted aggregation of expert and non-expert human rankings. On source-disjoint content across diverse codecs and bitrates, CodecArena attains state-of-the-art agreement with human judgments, surpassing both perceptual metrics and prior vision-language evaluators. Our contributions are:
\begin{itemize}
\item[\ding{113}] (1) We identify and analyze why perceptual metrics and existing quality models fail in the low and ultra-low bitrate coding regime, and cast codec evaluation as source-conditioned comparative reasoning in \textbf{CodecArena}.
\item[\ding{113}] (2) We propose \textbf{Facet-GRPO}, a facet-grounded preference learning scheme that combines holistic pairwise alignment with automatically derived facet-direction anchors, yielding hack-resistant and interpretable per-facet quality scores without requiring human per-facet labels.
\item[\ding{113}] (3) To address the lack of data for quality assessment of low and ultra-low bitrate video coding, we build a fully automatic pipeline that produces the 1{,}500-group \textbf{CodecArena-1K} training set and \textbf{CodecArena-Bench}, the first human-ranked benchmark for this task.
\item[\ding{113}] (4) On CodecArena-Bench, CodecArena aligns closely with human preference on source-disjoint content, surpassing both perceptual metrics and prior vision-language evaluators.
\end{itemize}

\section{Related Work}

\subsection{Evolution of Video Compression}

Video compression has long been driven by the rate-distortion trade-off. Traditional hybrid codecs, from H.264/AVC \citep{wiegand2003overview} and HEVC \citep{sullivan2012overview} to VVC \citep{bross2021overview} and AV1 \citep{han2021technical}, combine prediction, transform coding, quantization, in-loop filtering, and entropy coding to exploit spatial and temporal redundancy, yet at low bitrates information must still be discarded or approximated. Neural codecs instead replace hand-designed modules with learned components, from end-to-end predictive systems \citep{lu2019dvc,agustsson2020scale} to the DCVC family, which advances conditional coding through decoded feature contexts \citep{li2021deep,li2023neural,jia2025towards}.

In parallel, generative and perceptual codecs introduce adversarial losses and generative priors to produce visually plausible reconstructions under severe rate constraints, from generative image \citep{mentzer2020high} and video \citep{mentzer2021towards} compression to recent video-oriented methods such as GVC \citep{chen2025generative}, GLC-Video \citep{qi2025generative}, and GNVC-VD \citep{mao2025generative}. Diffusion priors push perceptual quality still higher at extremely low bitrates for both images \citep{xu2025picd,jia2025cod} and video \citep{xue2025s2vc,zheng2026gvc1d}. This shift from signal-level toward learned and perceptual reconstruction broadens compression artifacts beyond blocking, blur, and ringing; low-bitrate reconstruction may now also alter object identity, semantic consistency, fine structure, and temporal consistency.

\subsection{Quality Assessment for Video Coding}

Objective metrics remain the main protocol for comparing codecs. Full-reference metrics such as PSNR, SSIM \citep{wang2004image}, and MS-SSIM \citep{wang2003multiscale}, together with perceptual metrics such as LPIPS \citep{zhang2018unreasonable}, DISTS \citep{ding2022image}, and VMAF \citep{li2016toward}, improve correlation with human judgment. Nevertheless, they penalize visually acceptable changes while missing semantic or identity-level failures, and they are applied as feature distances rather than as source-conditioned reasoning over what content should be preserved.

More recently, multimodal large language models and vision-language models (VLMs) have moved beyond scalar scores toward interpretable, reasoning-based evaluation \citep{wu2024qbench,you2024depicting,ge2024lmm}, and reward models trained on large preference datasets capture fine-grained human preferences for generation \citep{he2024videoscore,xu2024visionreward,wang2025unifiedreward}. Casting quality understanding as reinforcement learning, Q-Insight \citep{li2025qinsight} and VQ-Insight \citep{zhang2025vqinsight} show that reward-driven reasoning yields generalizable evaluators. These methods, however, target natural, user-generated, or AI-generated content rather than codec-induced distortions, and do not compare a source video against multiple reconstructions from heterogeneous traditional, neural, and generative codecs. CodecArena is designed for exactly this comparative, source-conditioned setting.

\section{Methodology}

\subsection{Preliminaries}

Group Relative Policy Optimization (GRPO) \citep{shao2024deepseekmath} is a reinforcement learning framework for aligning large language and vision-language models. Unlike PPO \citep{schulman2017proximal}, which relies on a separate value network to estimate returns, GRPO removes the explicit critic and instead derives advantages from the relative quality of a group of responses sampled for the same query, reducing both compute overhead and the instability of a learned value function.
Given a query $q$, GRPO samples $N$ candidate responses $\{o_1, \dots, o_N\}$ from the old policy $\pi_{\theta_{\text{old}}}$, each assigned a scalar reward $r_i$ by task-specific reward functions. The relative advantage is obtained by normalizing the rewards within the group:
\begin{equation}
\hat{A}_i = \frac{r_i - \text{mean}(\{r_1, \dots, r_N\})}{\text{std}(\{r_1, \dots, r_N\})}.
\end{equation}
The policy is then updated to favor high-advantage responses under a clipped objective with a KL regularization term:
\begin{align}
\mathcal{J}_{GRPO}(\theta) &= \mathbb{E}_{[q \sim Q, o_{i} \sim \pi_{\theta_{\text{old}}}(o|q)]}\left\{
\min \left[
    \rho_i\hat{A}_i,  \right.\right. \\ \notag
&\left.\left.\operatorname{clip}(\rho_i, 1-\delta, 1+\delta)\hat{A}_i
\right] - \beta \cdot\mathbb{D}_{\mathrm{KL}}[\pi_\theta \| \pi_{\mathrm{ref}}]
\right\},
\end{align}
where $\rho_i = \pi_{\theta_{\text{new}}}(o_i \mid q) / \pi_{\theta_{\text{old}}}(o_i \mid q)$, $\delta$ bounds the per-step update, and $\beta$ weights the KL penalty against a frozen reference policy $\pi_{\text{ref}}$. As GRPO is agnostic to the reward form, its effectiveness hinges on the design of the reward functions.

\begin{figure}[!t]
    \centering
    \includegraphics[width=\columnwidth]{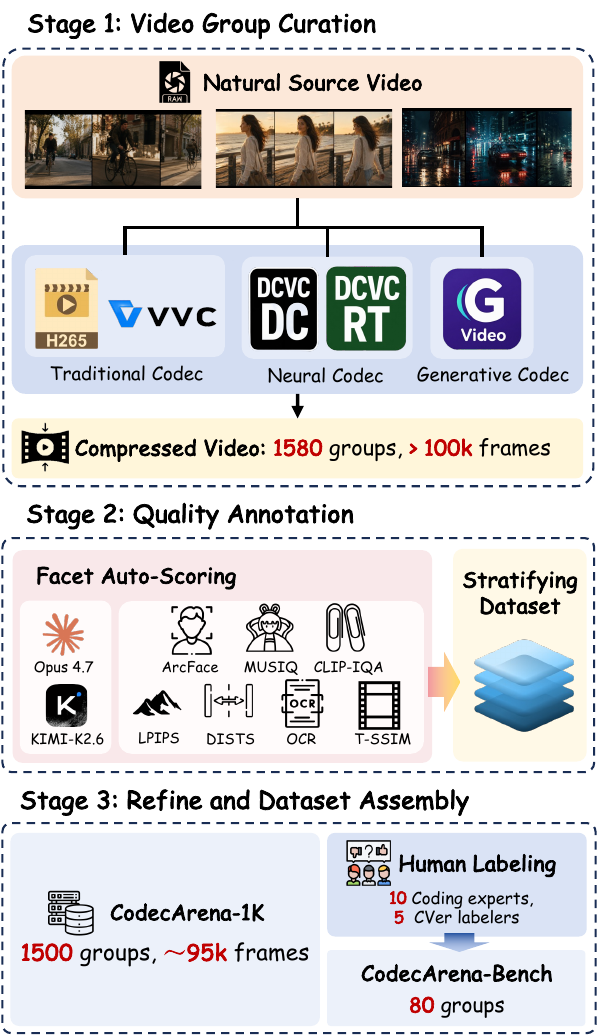}
    \caption{Our coding-oriented preference data curation pipeline. Stage 1 re-encodes natural source videos with traditional, neural, and generative codecs and curates the reconstructions into comparison groups. Stage 2 scores every group automatically by fusing the VLM-ensemble judgments with objective signals. Stage 3 assembles CodecArena-1K from the automatic labels and refines CodecArena-Bench with weighted human rankings.}
    \label{fig:curation}
\end{figure}

\subsection{Motivations of Our CodecArena}

At low and ultra-low bitrates, generative codecs synthesize visually sharp reconstructions that score well under LPIPS and DISTS yet corrupt semantically critical content, with drifting identities, illegible characters, and flicker across frames. Figure~\ref{fig:badcase} shows a concrete case in which both metrics prefer a generative reconstruction that has already lost the person's identity and made the on-screen text unreadable. Because they aggregate feature distances over the whole frame, these metrics overlook such localized failures; because they operate frame by frame, they also miss temporal incoherence. Many generative codecs also adopt an LPIPS-based term in their rate-distortion objective, so evaluating with the same insensitive metric lets a codec inflate its score without preserving the source. This motivates CodecArena, a framework that, conditioned on the source video, judges how faithfully each reconstruction preserves the original in semantics, structure, and temporal consistency.

Compression degradations also differ fundamentally from the synthetic or in-the-wild distortions studied by VLM-based quality assessment such as Q-Insight \citep{li2025qinsight}. They are shaped by rate constraints, prediction, and codec-specific tools, couple spatial distortion with temporal behavior, and are even less understood for generative coding, where part of the reconstruction is hallucinated. Reliable quality assessment for the coding scenario is therefore essential, which CodecArena is built to provide.

\subsection{Coding-oriented Preference Data Curation}

Existing video quality datasets target natural, user-generated, or AI-generated distortions and overlook codec-induced degradations, especially the hallucinated artifacts that generative coding produces at low bitrate; they also rarely provide source-conditioned labels over several reconstructions of the same content. To fill this gap, the three-stage pipeline in Figure~\ref{fig:curation} builds two resources: \textbf{CodecArena-1K}, a fully automatically labeled training corpus of 1{,}500 source-conditioned comparison groups, and \textbf{CodecArena-Bench}, a human-verified benchmark of 80 held-out source videos.

\begin{figure}[!t]
    \centering
    \includegraphics[width=\columnwidth]{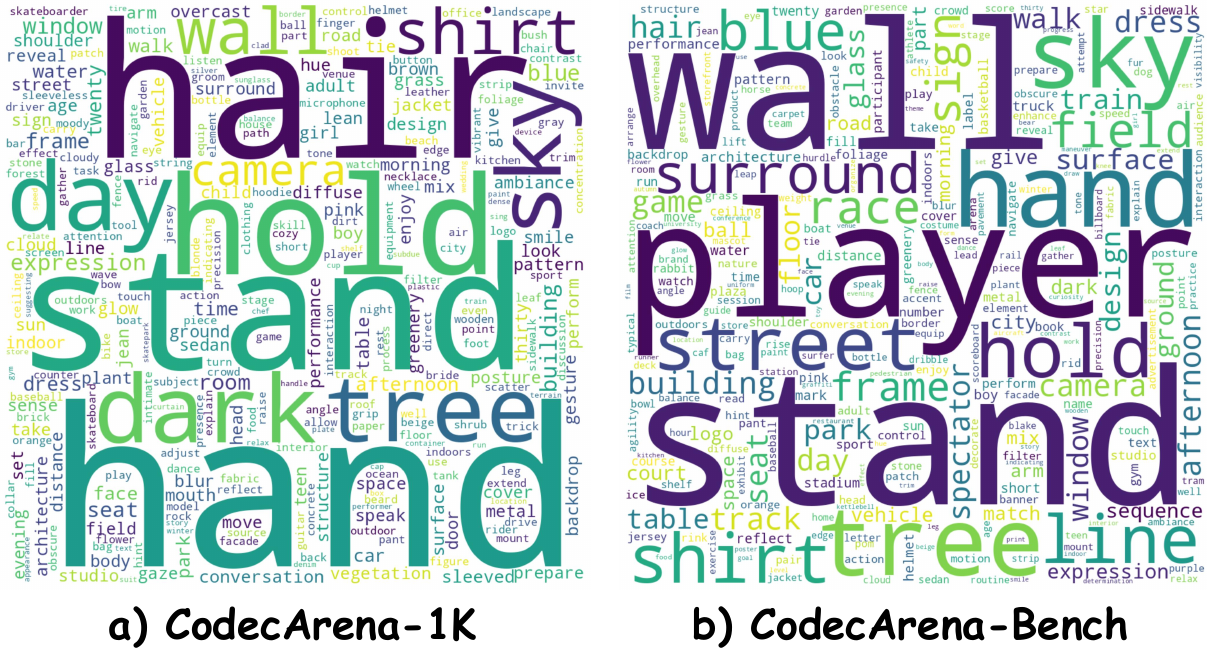}
    \caption{Word clouds of CodecArena-1K (a) and CodecArena-Bench (b). The two datasets share no source content, supporting fair evaluation on held-out content, while each spans diverse subjects, scenes, and objects.}
    \label{fig:wordcloud}
\end{figure}

\textbf{(1) Multi-codec re-encoding and group curation.} We collect natural source clips from five complementary datasets, Vimeo-90K \citep{xue2019video}, DAVIS~2019 \citep{caelles2019davis}, BVI-AOM \citep{nawala2024bviaom}, OpenVid \citep{nan2025openvid}, and Xiph.org test sequences \citep{xiph_media}, spanning content that stresses faces, text, motion, and texture, with training and benchmark clips drawn from disjoint sources to evaluate generalization to held-out content. Each source is re-encoded by all three codec families, HEVC \citep{sullivan2012overview} and VVC \citep{bross2021overview}, the neural DCVC-DC \citep{li2023neural} and DCVC-RT \citep{jia2025towards}, and the generative GLC-Video \citep{qi2025generative}. To our knowledge GLC‑Video is the only generative video codec with publicly released implementation, as the authors provide an official codebase with both image and video compression support. Reconstructions from these codecs form comparison groups that balance content, codec family and rate, so that evaluations hinge on subtle semantic and temporal differences rather than gross distortion.

\textbf{(2) Quality annotation.} Every group is scored automatically. A vision-language judge, an ensemble of Kimi-K2.6 \citep{kimi2026k2} and Opus-4.7 \citep{anthropic2026claude} for reliability, compares the reconstructions pairwise over seven-frame filmstrips and rates each side from 0 to 1 along five facets, identity, objects, text, texture, and temporal consistency, under a rubric that penalizes hallucinated or identity-shifting content over honest blur. We fuse this VLM score with seven objective signals spanning face identity, full- and no-reference quality, text legibility, and temporal consistency into a single quality score, normalizing per group before a weighted average that also enforces the within-codec rate monotonicity the VLM alone does not guarantee.

\textbf{(3) Grading and dataset assembly.} Ranking each group by this score into ordinal grades yields \textbf{CodecArena-1K}, the 1{,}500-group corpus that supervises CodecArena without manual annotation. For \textbf{CodecArena-Bench}, 15 annotators, 10 video-coding experts and 5 non-domain viewers, independently re-rank every group, and we aggregate their orderings under an expertise-weighted scheme into human labels over 80 videos spanning diverse scenes, resolutions, and codecs at medium-low to ultra-low bitrate. As the word clouds in Figure~\ref{fig:wordcloud} confirm, both datasets cover diverse subjects, scenes, and objects despite disjoint sources, so evaluation on held-out content remains representative. The complete pipeline, the exact objective signals, and the fusion weights are given in the supplementary.

\subsection{Facet-GRPO Preference Learning}

\begin{figure*}[t]
    \centering
    \includegraphics[width=\textwidth]{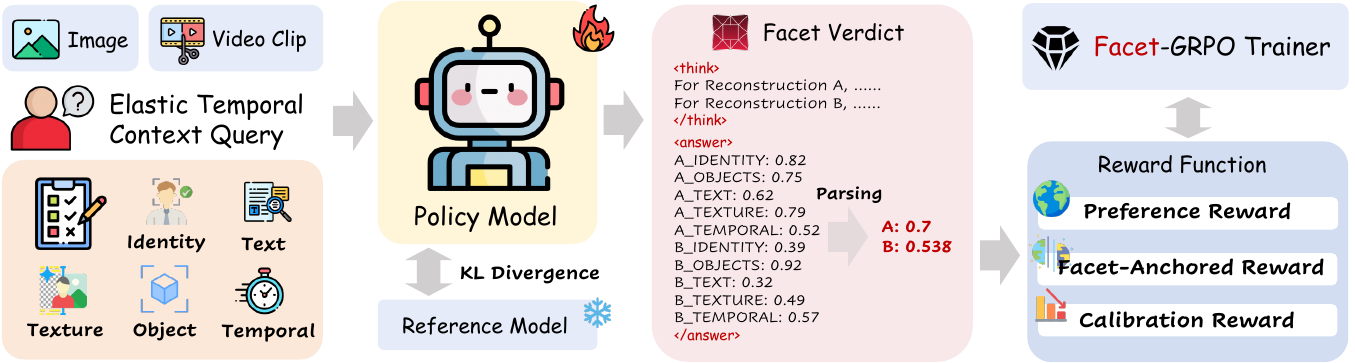}
    \caption{Overview of CodecArena's Facet-GRPO training. An \emph{Elastic Temporal Context Query} presents the reference and two reconstructions at a variable temporal granularity, from a single still image to a multi-frame clip, together with the quality facets to be judged. The trainable policy reasons inside \texttt{<think>} and emits a per-facet \emph{Facet Verdict} inside \texttt{<answer>}, regularized toward a frozen reference model by a KL term. Parsing the verdict yields the side-level qualities that the Facet-GRPO trainer scores with the Preference, Facet-Anchored, and Calibration rewards.}
    \label{fig:framework}
\end{figure*}

We post-train the vision-language judge with Facet-GRPO, a group-relative scheme tailored to source-conditioned codec comparison, summarized in Figure~\ref{fig:framework}. A query $q=(R,A,B)$ presents a reference video $R$ and two reconstructions $A$ and $B$ as filmstrips, and the policy response $o$ ends with a structured verdict that assigns a score $s_X^{k}(o)\in[0,1]$ to each side $X\in\{A,B\}$ and facet $k\in\mathcal{F}=\{\text{identity, objects, text, texture, temporal}\}$, with facets absent from the reference marked NA. The side-level quality $\bar{s}_X(o)$ is the mean over present facets and the preferred side is $w(o)=\arg\max_X\bar{s}_X(o)$. The curated data supplies three supervision signals per comparison, namely Bradley--Terry latent qualities $b_A,b_B$ encoding the holistic preference, an objective composite quality $y_X\in[0,1]$ per side, and per-facet objective signals $y_X^{k}$. The central risk in this setting is reward hacking. Since $\bar{s}_X$ averages over facets, a policy can win a comparison by inflating one salient facet, such as synthesized texture, while a face or a caption is corrupted, mirroring the very failure of LPIPS-style feedback. Facet-GRPO therefore instantiates the reward $r_i$ of Eq.~(1) as complementary terms that constrain the verdict at both the holistic and the facet level, optimized under an elastic temporal context.

\textbf{Reasoning and format rewards.} Following the think-then-answer convention of Q-Insight \citep{li2025qinsight}, the policy reasons inside \texttt{<think>} tags and emits the score lines inside \texttt{<answer>} tags. A format term $r_{\mathrm{fmt}}$ grants a small constant when the facet scores of both sides are well formed, and a reasoning term $r_{\mathrm{think}}$ grants full credit only when a substantive reasoning block precedes a well-formed answer, reduced credit when reasoning is present but the answer is malformed, and zero when reasoning is absent. Without $r_{\mathrm{think}}$, GRPO gradually collapses the chain-of-thought into score-only outputs; with it, the verdict remains grounded in explicit visual evidence.

\textbf{Preference and calibration rewards.} The \emph{Preference Reward} aligns the model's choice with the curated ranking through the Bradley--Terry probability
\begin{equation}
r_{\mathrm{pref}}(o)=\sigma\!\big(b_{w(o)}-b_{l(o)}\big),
\end{equation}
where $l(o)$ denotes the rejected side and $\sigma$ the logistic function; a confident pair thus rewards a correct call near $1$ and a wrong one near $0$. The \emph{Calibration Reward} matches the magnitude of the side-level quality to the objective composite,
\begin{equation}
r_{\mathrm{reg}}(o)=1-\tfrac{1}{2}\big(\,|\bar{s}_A(o)-y_A|+|\bar{s}_B(o)-y_B|\,\big).
\end{equation}
Together they fix the ordering and the scale of the verdict.

\textbf{Facet-anchored reward.} The preference and calibration terms leave the facet decomposition under-determined, precisely the loophole that hacking exploits. The \emph{Facet-Anchored Reward} closes it. Let $\mathcal{D}(q)=\{k:|y_A^{k}-y_B^{k}|>\epsilon\}$ collect the facets on which the objective signals separate the two sides unambiguously, with direction $p^{k}=\operatorname{sign}(y_A^{k}-y_B^{k})$. We reward agreement with every such direction,
\begin{equation}
r_{\mathrm{facet}}(o)=\frac{1}{|\mathcal{D}(q)|}\sum_{k\in\mathcal{D}(q)}\sigma\!\big(\gamma\,p^{k}\,(s_A^{k}(o)-s_B^{k}(o))\big),
\end{equation}
where $\gamma$ sharpens the per-facet margin and a comparable facet left unscored contributes zero. Credit is shared across all separated facets, such that no single sub-score can carry the comparison. This term injects no human per-facet target value, only the direction in which the automatically derived signals already order the pair, so calibrated per-facet scoring emerges without human per-facet annotation.

\textbf{Elastic temporal context query.} Rather than fixing the input format, we pose each comparison as an elastic temporal context query: at every training step we draw a frame count $K\in\{1,3,5,7\}$ and render the whole batch, reference and reconstructions alike, at that granularity. For $K>1$ the query is a short clip and all five facets apply; for $K=1$ it degenerates to a single decoded still, an image-level assessment in which the temporal facet is declared not scorable and removed from $\mathcal{D}(q)$, while the remaining rewards operate unchanged. Treating temporal context as an elastic training variable rather than a fixed format yields a single judge that assesses codec quality from one still frame to a full clip.

\textbf{Overall objective.} The terms combine into
\begin{equation}
r_i=r_{\mathrm{fmt}}+r_{\mathrm{pref}}+r_{\mathrm{reg}}+\lambda_{f}\,r_{\mathrm{facet}}+\lambda_{t}\,r_{\mathrm{think}},
\end{equation}
where $\lambda_{f}$ controls the facet-direction anchors and $\lambda_{t}$ controls the reasoning bonus. The reward enters the group-relative advantage of Eq.~(1) and is optimized with Eq.~(2) over $N$ sampled verdicts, requiring neither a value network nor a learned reward model. Unlike Q-Insight, which regresses a holistic image score with auxiliary degradation labels, or VQ-Insight \citep{zhang2025vqinsight}, which regresses dense opinion scores for a single generated video, Facet-GRPO is comparative and source-conditioned. It grounds decomposition through directional, automatically derived facet signals plus holistic preferences, yielding hack-resistant per-facet scores without human per-facet labels.

\section{Experiments}

\subsection{Implementation Details}

We instantiate CodecArena on Qwen2.5-VL-7B-Instruct \citep{bai2025qwen25vl} and Qwen3-VL-8B-Instruct \citep{qwen2025qwen3vl}. The 1{,}500 groups of CodecArena-1K are split into 1{,}050 training and 450 held-out validation groups, and the judges are evaluated on the source-disjoint CodecArena-Bench. We set $N{=}8$, $\beta{=}1\times10^{-3}$, $\lambda_f{=}0.6$, and $\lambda_t{=}0.2$, and fine-tune each model for 2 epochs on 8 NVIDIA A800 GPUs with learning rate $1\times10^{-6}$ and effective batch size 128. Each query samples $K\in\{1,3,5,7\}$.

\subsection{Quantitative Results}

No established benchmark exists for this regime, so we evaluate on the human-labeled, source-disjoint \textbf{CodecArena-Bench}. Each protocol is tested at $K{=}1$ (image), $K{=}7$ (trained clip length), and $K{=}14$ (extrapolation). Baselines include PSNR, SSIM \citep{wang2004image}, LPIPS \citep{zhang2018unreasonable}, DISTS \citep{ding2022image}, VMAF \citep{li2016toward}, Q-ReAlign-Pro-9B \citep{qrealignpro}, DeQA-Score \citep{you2025deqa}, Q-Insight \citep{li2025qinsight}, and VQ-Insight \citep{zhang2025vqinsight}. We additionally report a VLM Ensemble, the arithmetic mean of Opus-4.7~\cite{anthropic2026claude} and Kimi-K2.6~\cite{kimi2026k2} scores. Q-Insight and VQ-Insight are developed on Qwen2.5-VL-7B \citep{bai2025qwen25vl}, while Q-ReAlign-Pro-9B is an improved Q-Align \citep{wu2024qalign} built on Qwen3.5-9B \citep{qwen35blog}.

\begin{table}[t]
\centering
\scriptsize
\setlength{\tabcolsep}{1pt}
\caption{\textbf{Score-based ranking on CodecArena-Bench}. Best/second-best are \textbf{bold}/\underline{underlined}.}
\label{tab:scoring}
\begin{tabular*}{\columnwidth}{@{\extracolsep{\fill}}lcccccc}
\toprule
 & \multicolumn{2}{c}{$K{=}1$} & \multicolumn{2}{c}{$K{=}7$} & \multicolumn{2}{c}{$K{=}14$} \\
\cmidrule(lr){2-3}\cmidrule(lr){4-5}\cmidrule(lr){6-7}
Method & SRCC$\uparrow$ & PLCC$\uparrow$ & SRCC$\uparrow$ & PLCC$\uparrow$ & SRCC$\uparrow$ & PLCC$\uparrow$ \\
\midrule
\rowcolor{gray!18}\multicolumn{7}{l}{\textit{Score-based metrics}}\\
PSNR                                  & 0.687 & 0.817 & 0.808 & 0.876 & 0.813 & 0.876 \\
SSIM~\citep{wang2004image}            & 0.753 & 0.833 & 0.829 & 0.892 & 0.841 & 0.891 \\
LPIPS~\citep{zhang2018unreasonable}   & 0.798 & 0.817 & 0.723 & 0.758 & 0.748 & 0.763 \\
DISTS~\citep{ding2022image}           & 0.661 & 0.722 & 0.658 & 0.729 & 0.669 & 0.724 \\
VMAF~\citep{li2016toward}             & 0.782 & 0.853 & 0.876 & \underline{0.919} & 0.878 & \underline{0.917} \\
\midrule
\rowcolor{gray!18}\multicolumn{7}{l}{\textit{Vision-language evaluators}}\\
VLM Ensemble                          & 0.798 & 0.799 & 0.794 & 0.798 & 0.753 & 0.766 \\
Q-ReAlign-Pro-9B~\citep{qrealignpro}  & 0.636 & 0.675 & 0.642 & 0.679 & 0.691 & 0.687 \\
DeQA-Score~\citep{you2025deqa}        & 0.511 & 0.650 & 0.483 & 0.647 & 0.461 & 0.632 \\
Q-Insight~\citep{li2025qinsight}      & 0.587 & 0.650 & 0.589 & 0.651 & 0.614 & 0.659 \\
VQ-Insight~\citep{zhang2025vqinsight} & 0.663 & 0.681 & 0.573 & 0.639 & 0.672 & 0.674 \\
\textbf{CodecArena} (Qwen2.5-VL-7B)   & \underline{0.871} & \underline{0.906} & \textbf{0.902} & 0.912 & \textbf{0.906} & 0.908 \\
\quad $\Delta$ vs. Qwen2.5-VL-7B-Instruct & +0.235 & +0.228 & +0.407 & +0.365 & +0.456 & +0.413 \\
\textbf{CodecArena} (Qwen3-VL-8B)     & \textbf{0.877} & \textbf{0.909} & \underline{0.895} & \textbf{0.923} & \underline{0.894} & \textbf{0.924} \\
\quad $\Delta$ vs. Qwen3-VL-8B-Instruct & +0.161 & +0.185 & +0.190 & +0.198 & +0.199 & +0.228 \\
\bottomrule
\end{tabular*}
\end{table}

\textbf{Score-based ranking.} Each model scores one reconstruction against the reference, inducing a within-group ranking evaluated by SRCC and PLCC. Table~\ref{tab:scoring} shows that VMAF is the strongest non-learned metric and remains competitive on PLCC, but the two CodecArena variants occupy the top SRCC positions at every granularity. LPIPS and DISTS fall behind PSNR and VMAF on SRCC, confirming that feature similarity diverges from human preference here. CodecArena also beats all language baselines, including the larger Q-ReAlign-Pro-9B, and maintains accuracy at $K{=}14$ despite training only with $K{\le}7$.

\begin{table}[t]
\centering
\footnotesize
\setlength{\tabcolsep}{2pt}
\caption{\textbf{Pairwise preference accuracy on CodecArena-Bench}. Best/second-best are \textbf{bold}/\underline{underlined}.}
\label{tab:pairwise}
\begin{tabular*}{\columnwidth}{@{\extracolsep{\fill}}lccc}
\toprule
 & \multicolumn{3}{c}{Accuracy (\%)\,$\uparrow$} \\
\cmidrule(lr){2-4}
Method & $K{=}1$ & $K{=}7$ & $K{=}14$ \\
\midrule
\rowcolor{gray!18}\multicolumn{4}{l}{\textit{Score-based metrics}}\\
PSNR                                  & 80.8 & 85.8 & 86.7 \\
SSIM~\citep{wang2004image}            & 83.3 & 87.9 & 88.9 \\
LPIPS~\citep{zhang2018unreasonable}   & 82.9 & 79.6 & 78.9 \\
DISTS~\citep{ding2022image}           & 73.8 & 75.0 & 76.1 \\
VMAF~\citep{li2016toward}             & 85.8 & 89.2 & 90.0 \\
\midrule
\rowcolor{gray!18}\multicolumn{4}{l}{\textit{Vision-language evaluators}}\\
VLM Ensemble                          & 85.2 & 80.9 & 83.6 \\
Q-ReAlign-Pro-9B~\citep{qrealignpro}  & 78.3 & 77.5 & 81.1 \\
DeQA-Score~\citep{you2025deqa}        & 73.4 & 72.1 & 73.9 \\
Q-Insight~\citep{li2025qinsight}      & 80.9 & 76.5 & 75.6 \\
VQ-Insight~\citep{zhang2025vqinsight} & 81.8 & 79.0 & 83.4 \\
\textbf{CodecArena} (Qwen2.5-VL-7B)   & \underline{89.5} & \underline{92.7} & \underline{92.3} \\
\textbf{CodecArena} (Qwen3-VL-8B)     & \textbf{93.7} & \textbf{94.9} & \textbf{93.0} \\
\bottomrule
\end{tabular*}
\end{table}

\textbf{Pairwise preference.} Each model selects the better reconstruction from a randomly ordered A/B pair. Table~\ref{tab:pairwise} shows that CodecArena is first and second at every granularity; the Qwen3-VL-8B model reaches $93.7\%$, $94.9\%$, and $93.0\%$ accuracy from $K{=}1$ to $K{=}14$. VMAF's strong PLCC does not translate into pairwise choice, and the language baselines collapse close comparisons to a single appearance impression, whereas CodecArena weighs per-facet differences.

\subsection{Ablation Study}

\begin{table}[t]
\centering
\scriptsize
\setlength{\tabcolsep}{1pt}
\caption{\textbf{Ablation of Facet-GRPO components}, added cumulatively to Qwen3-VL-8B. The full model uses $\lambda_f{=}0.6$; metrics follow Table~\ref{tab:scoring}.}
\label{tab:ablation}
\begin{tabular*}{\columnwidth}{@{\extracolsep{\fill}}lcccccc}
\toprule
 & \multicolumn{2}{c}{$K{=}1$} & \multicolumn{2}{c}{$K{=}7$} & \multicolumn{2}{c}{$K{=}14$} \\
\cmidrule(lr){2-3}\cmidrule(lr){4-5}\cmidrule(lr){6-7}
Configuration & SRCC$\uparrow$ & PLCC$\uparrow$ & SRCC$\uparrow$ & PLCC$\uparrow$ & SRCC$\uparrow$ & PLCC$\uparrow$ \\
\midrule
Qwen3-VL-8B-Instruct & 0.716 & 0.724 & 0.705 & 0.725 & 0.695 & 0.696 \\
\quad +\,Preference reward     & 0.770 & 0.795 & 0.789 & 0.831 & 0.782 & 0.819 \\
\quad +\,Elastic context       & 0.770 & 0.822 & 0.841 & 0.861 & 0.794 & 0.832 \\
\quad +\,Calibration reward    & 0.844 & 0.872 & 0.894 & 0.905 & 0.884 & 0.897 \\
\textbf{\quad +\,Facet-anchored (Ours)} & \textbf{0.877} & \textbf{0.909} & \textbf{0.895} & \textbf{0.923} & \textbf{0.894} & \textbf{0.924} \\
\bottomrule
\end{tabular*}
\end{table}

\begin{figure*}[t]
    \centering
    \includegraphics[width=\textwidth]{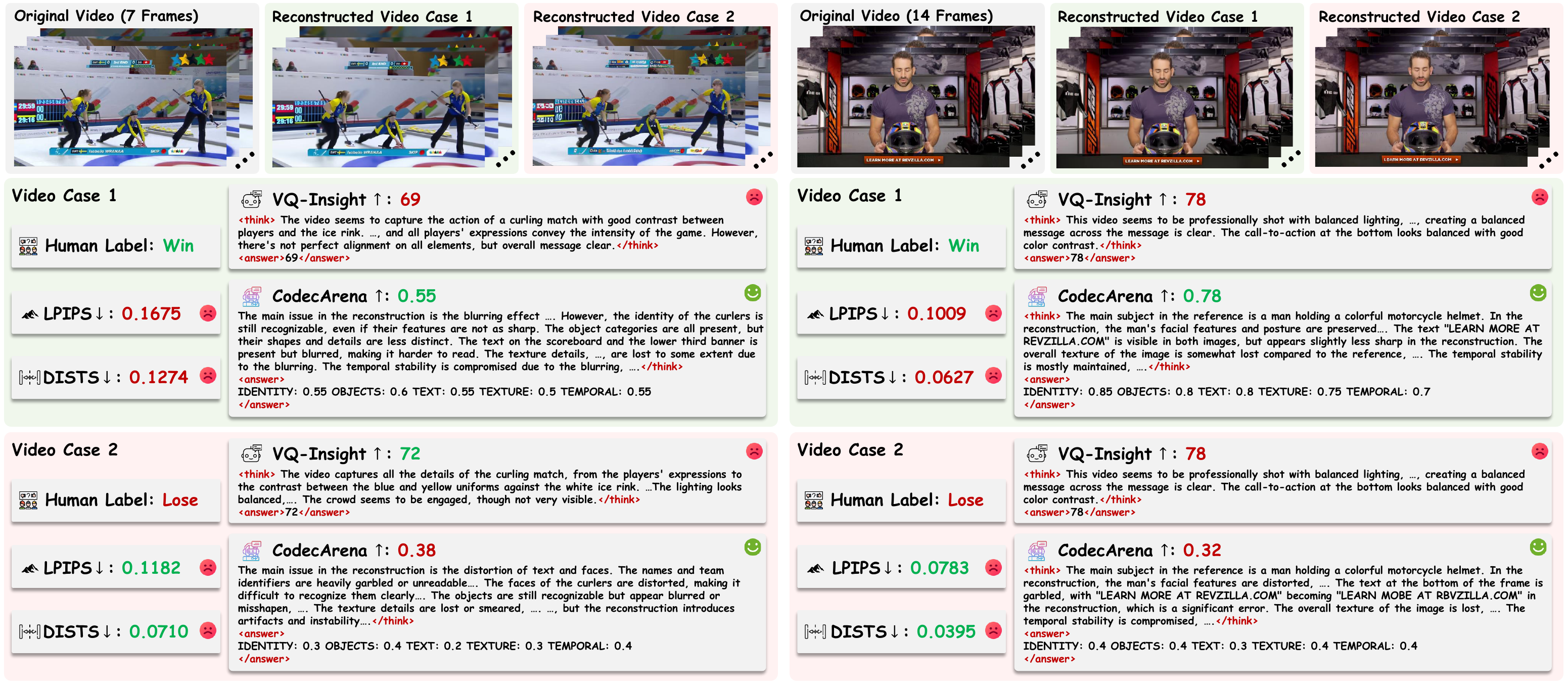}
    \caption{\textbf{Single-stimulus scoring on a $7$-frame (left) and $14$-frame (right) clip.} CodecArena aligns with the human preference, while LPIPS, DISTS, and VQ-Insight do not.}
    \label{fig:qual_video}
\end{figure*}

\begin{figure}[!t]
    \centering
    \includegraphics[width=\columnwidth]{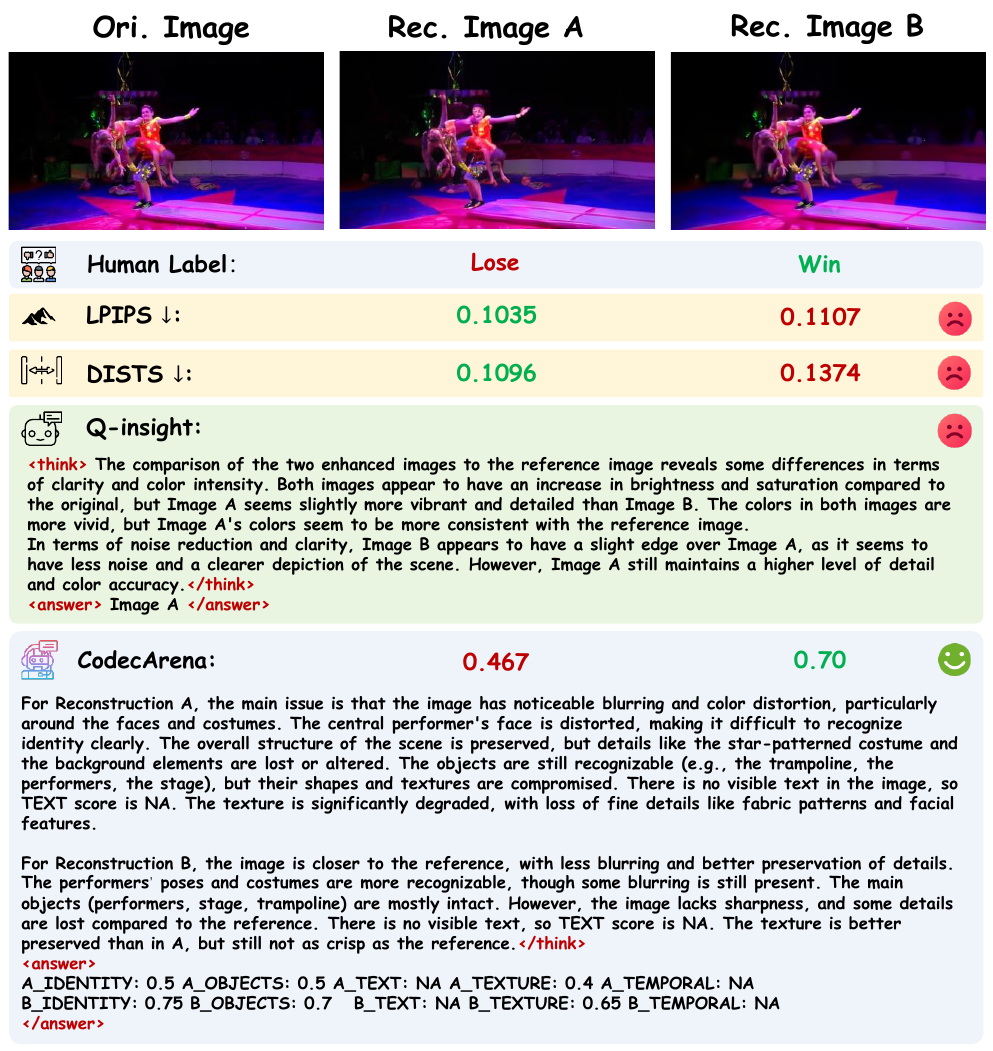}
    \caption{\textbf{Pairwise preference at $K{=}1$.} Although humans prefer B, only CodecArena aligns with this preference, while LPIPS, DISTS, and Q-Insight favor A.}
    \label{fig:qual_image}
\end{figure}

We ablate Facet-GRPO by cumulatively adding each reward component to the native Qwen3-VL-8B backbone in Table~\ref{tab:ablation}. Preference supervision alone is insufficient, as it improves coarse ranking but still leaves the model vulnerable to under-specified reasoning and unstable facet scores. Elastic context, calibration, and facet anchoring each improve ranking, with calibration giving the largest jump and the final facet anchor further regularizing sub-scores. We treat this as evidence that automatic facet directions provide complementary structure beyond holistic preference supervision, while the supplementary hard-case audit directly measures their effect on facet-level reward hacking. \textbf{More ablations are present in Appendix.}

\subsection{Qualitative Results}

Figures~\ref{fig:qual_video} and~\ref{fig:qual_image} cover a clip and a single image with the same judge. LPIPS, DISTS, VMAF, and prior VLM evaluators fold these errors into one appearance score and may prefer the lower-distortion but unfaithful reconstruction, while CodecArena's facet scores remain grounded by objective signals. \textbf{More qualitative results are provided in the appendix.}

These cases clarify why the pairwise gains in Table~\ref{tab:pairwise} are larger than the score-correlation gains in Table~\ref{tab:scoring}. Codec decisions often hinge on a localized but decisive failure: one reconstruction may be slightly sharper overall while changing a face, corrupting a caption, or introducing temporal drift. A single scalar metric can still correlate with broad quality trends and yet choose the wrong side in exactly these close comparisons. CodecArena instead compares the source and both reconstructions facet by facet, so the final preference is tied to the content that humans use to reject unfaithful generative outputs. The protocol also makes the reported gains conservative. Each benchmark group compares reconstructions of the same source, so color, composition, and semantic content are shared and the decision turns on what each codec preserves or fabricates. The held-out sources are disjoint from CodecArena-1K, and the $K{=}14$ setting further tests a temporal context never seen during training. The remaining errors are therefore not dominated by easy cases such as obvious blur or severe blocking; they arise from fine trade-offs between honest softness and plausible but wrong detail. This is precisely where automatic metrics are most likely to reward generative sharpness, and where a source-conditioned facet judge is most useful.

\section{Conclusion}

We presented CodecArena, a source-conditioned VLM judge for low- and ultra-low-bitrate codec assessment. Facet-GRPO learns from holistic preferences while grounding fine-grained scores with automatic facet-direction anchors, avoiding human per-facet labels. This design targets the evaluation gap created by generative codecs: a reconstruction can look sharper and still be less faithful to the source. By separating identity, objects, text, texture, and temporal consistency, CodecArena makes such trade-offs explicit rather than hiding them in one appearance score. This also changes what codec progress means in the generative regime: quality is not only smoother pixels or lower feature distance, but preservation of the source information a viewer relies on. Together with CodecArena-1K and the 80-group human-ranked CodecArena-Bench, it aligns with human rankings more closely than perceptual metrics and prior VLM evaluators, especially when plausible reconstructions alter identity, text, or temporal consistency. We hope this benchmark and judge encourage codec research to report semantic fidelity alongside rate-distortion efficiency, making future codec comparisons more faithful to the failures viewers actually notice and less dependent on metrics that generative models can exploit. This matters as learned decoders increasingly synthesize missing content rather than merely compress transmitted detail.

\bibliography{aaai2027}



\end{document}